\pdfoutput=1

\documentclass[11pt]{article}

\usepackage[preprint]{acl}

\usepackage{times}
\usepackage{latexsym}

\usepackage[T1]{fontenc}

\usepackage[utf8]{inputenc}

\usepackage{microtype}

\usepackage{inconsolata}

\usepackage{graphicx}

\usepackage{hyperref}
\usepackage{xcolor}
\usepackage{tcolorbox}
\usepackage{amsmath}
\usepackage{amsfonts}
\usepackage{nicefrac}
\usepackage{multirow}
\usepackage{svg}
\tcbuselibrary{breakable, skins}
\usepackage{booktabs}
\usepackage{url}
\usepackage{xspace}
\usepackage{float}
\usepackage{textcomp}

\newcommand{\data}{\textbf{s1K-mix}}
\newcommand{\model}{\textbf{s1-mix-32B}}

\title{Long-Short Chain-of-Thought Mixture Supervised Fine-Tuning Eliciting Efficient Reasoning in Large Language Models}

\author{
 \textbf{Bin Yu\textsuperscript{1,3}},
 \textbf{Hang Yuan\textsuperscript{2,3}},
 \textbf{Haotian Li\textsuperscript{1,4}},
 \textbf{Xueyin Xu\textsuperscript{3,4}},
\\
 \textbf{Yuliang Wei\textsuperscript{1}},
 \textbf{Bailing Wang\textsuperscript{1}},
 \textbf{Weizhen Qi\textsuperscript{3,4}},
 \textbf{Kai Chen\textsuperscript{3,4}},
\\
\\
 \textsuperscript{1}Harbin Institute of Technology,
 \textsuperscript{2}East China Normal University,
 \\
 \textsuperscript{3}Zhongguancun Academy,
 \textsuperscript{4}Zhongguancun Institute of Artificial Intelligence
\\
 \small{
   \textbf{Correspondence:} \href{mailto:weizhenqi@zgci.ac.cn}{weizhenqi@zgci.ac.cn}
 }
}

\begin{document}
\maketitle

\begin{abstract}
\label{sec:abstract}
    
Recent advances in large language models have demonstrated that Supervised Fine-Tuning (SFT) with Chain-of-Thought (CoT) reasoning data distilled from large reasoning models (e.g., DeepSeek R1) can effectively transfer reasoning capabilities to non-reasoning models. However, models fine-tuned with this approach inherit the "overthinking" problem from teacher models, producing verbose and redundant reasoning chains during inference. To address this challenge, we propose \textbf{L}ong-\textbf{S}hort Chain-of-Thought \textbf{Mixture} \textbf{S}upervised \textbf{F}ine-\textbf{T}uning (\textbf{LS-Mixture SFT}), which combines the long CoT reasoning dataset with their short counterparts obtained through structure-preserved rewriting. Our experiments demonstrate that models trained with the LS-Mixture SFT method achieved an average accuracy improvement of 2.3\% across various benchmarks  compared to those trained with standard SFT. Furthermore, this approach substantially reduced the model response length by approximately 47.61\%. This work offers an approach to endow non-reasoning models with reasoning capabilities through supervised fine-tuning while avoiding the inherent overthinking problems inherited from teacher models, thereby enabling efficient reasoning in the fine-tuned models.


\end{abstract}

\section{Introduction}

\begin{figure*}[!ht]
    \centering
    \includegraphics[trim=0 12 0 16, clip, scale=0.85]{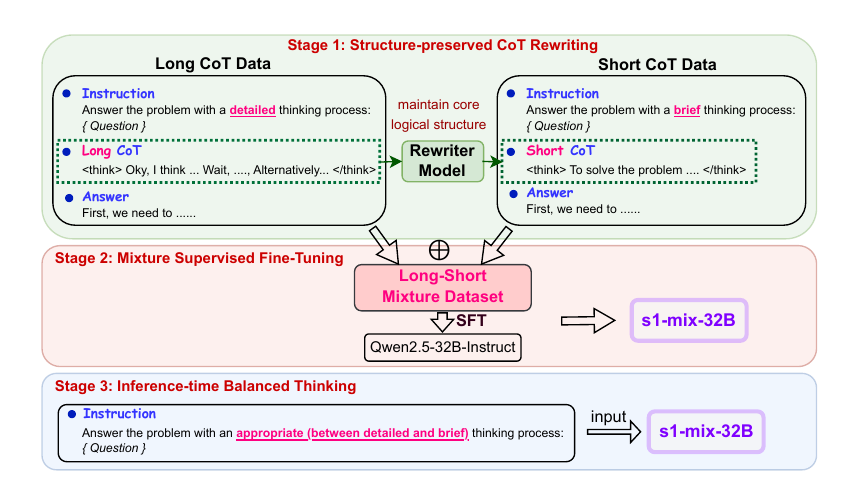}
    \caption{Overview of \textbf{LS-Mixture SFT}. This method consists of three stages: \textbf{1) Structure-preserved CoT Rewriting}: A LLM is used to rewrite the long CoT trajectories into short ones while preserving the core structure. \textbf{2) Mixture Supervised Fine-Tuning}: Non-reasoning LLM is been supervised fine-tuned on mixture datasets. \textbf{3) Inference-time Balanced Thinking}: The fine-tuned model is designed to employ a balanced thinking mode that lies between detailed and brief when generating reasoning responses to queries.}
    \label{fig:arch}
\end{figure*}

The emergence of large reasoning models (LRMs) \citep{TowardsReasoningEra-2503}, such as DeepSeek R1 \citep{DeepSeek-R1} and OpenAI o1 \citep{OpenAI-o1}, have demonstrated remarkable reasoning abilities in complex tasks by generating explicit chain-of-thoughts (CoT) \citep{CoT-2022}  closed by special tokens (\texttt{<think>} and \texttt{</think>}) about a question before arriving at the final answer. Recent works \citep{O1ReplicationJourney-part2, ReproductionReport-2412} have shown that advanced reasoning abilities can be transferred from LRMs to non-reasoning large language models (LLMs) through supervised fine-tuning (SFT) on high-quality CoT reasoning data distilled from LRM \citep{DeepSeek-R1, s1-2501}. 

Existing open-source efforts, such as s1 \citep{s1-2501}, Sky-T1 \citep{SkyT1-2501} and LIMO \citep{LIMO-2502}, have demonstrated that non-reasoning LLMs as student models can be effectively transformed into reasoning-capable models through supervised fine-tuning on long CoT trajectories distilled from LRMs as teacher models. Although training on distilled datasets successfully elicits reasoning abilities in foundation models, it also causes these models to inherit the inherent overthinking problem \citep{OverthinkingOfO1-2412} of the original LRM \citep{StopOverthinkingSurvey-2303, HarnessingReasoningEconomy-2503}. Several recent studies have sought to address the overthinking problem of LRM during training and inference from the perspectives of reinforcement learning and inference-time optimization, aiming to achieve efficient reasoning. However, there remains a lack of research on how to prevent student models from inheriting the overthinking issue of teacher models during the distillation stage. Thus, we propose the problem: "\textbf{how can data distillation and supervised fine-tuning be leveraged to elicit more \textit{efficient} reasoning abilities in non-reasoning models}—specifically, enabling them to avoid inheriting the overthinking problem from teacher models?".

In this paper, we propose a novel solution to this problem: Long-Short Chain-of-Thought Mixture Supervised Fine-Tuning (\textbf{LS-Mixture SFT}). Our approach first performs structure-preserved rewriting of the reasoning paths in the dataset with long CoT trajectories distilled from LRM, resulting in a corresponding dataset with short CoT reasoning paths. We then construct a mixture of both long and short CoT reasoning datasets, and use it to perform supervised fine-tuning on the student model. This mixture allows student models to learn both comprehensive reasoning patterns and efficient reasoning shortcuts, resulting in models that can generate more efficient reasoning during inference without sacrificing accuracy. Our approach can directly reuse existing long CoT reasoning datasets without incurring the substantial costs associated with additional data distillation. Specifically, we created a mixed dataset of long and short reasoning chains, \data{}, based on the existing s1K-1.1 dataset \citep{s1-2501}, and utilized this mixed dataset to train the Qwen2.5-32B-Instruct model, resulting in our model \model{}.

Our extensive experiments across three challenging reasoning benchmarks validate the effectiveness of the LS-Mixture SFT approach. The experimental results demonstrate that our s1-mix-32B model achieves higher accuracy on MATH500, AIME24, and GPQA benchmarks (improvements of 2.2\%, 6.7\%, and 2\%, respectively) compared to models trained solely on long-chain reasoning data, while significantly reducing average response length (by 47.61\% on average). Ablation studies further confirm the importance of our proposed structure-preserved CoT rewriting strategy and the advantages of the long-short chain mixture training method in balancing reasoning efficiency and accuracy. These findings indicate that LS-Mixture SFT not only effectively elicits reasoning capabilities in non-reasoning models but also successfully avoids the overthinking problem inherited from existing LRMs, providing an effective approach for training more efficient reasoning models.

Our contributions can be summarized as follows:

\begin{itemize}
    \item We propose a novel method for transforming long chain-of-thought trajectories into their short counterparts: Structure-preserved CoT Rewriting, which is designed to rewrite reasoning paths while preserving the core structure, thereby preventing overly liberal rewriting that might cause models to lose crucial "aha moments" ability during training.
    \item We introduce LS-Mixture SFT, a novel fine-tuning approach that mix long and short reasoning dataset to elicits efficient reasoning in large language models.
    \item Based on these methods, we build a new mixture dataset \data{} and a fine-tuned model \model{} released on HuggingFace.
    \item Through extensive experiments, we demonstrate that our approach significantly reduces model response length during inference while improving task performance.
    \item During our experiments, we observed an interesting phenomenon: the fine-tuned model's ability to success in balanced thinking was not explicitly trained but rather emerged as a natural consequence of training on a mixture dataset comprising both long-chain and short-chain reasoning examples.
\end{itemize}

Our code, model, and data are open-sourced at \href{https://github.com/ZGCA-AI4Edu/LS-Mixture}{GitHub} and \href{https://huggingface.co/datasets/VLyb/s1K-mix}{HuggingFace}.


\section{Methodology}  

In this section, we introduce Long-Short Chain-of-Thought Mixture Supervised Fine-Tuning (\textbf{LS-Mixture SFT}), our novel approach for efficiently transferring reasoning capabilities from LRMs to non-reasoning LLMs. The key insight of our method is that not all reasoning steps contribute equally to the final solution—many tokens in verbose reasoning chains are redundant. By leveraging LLMs as rewriter model that preserve the core reasoning structure while eliminating redundancy, we create a complementary dataset of short reasoning examples. These shortened trajectories maintain the core structure and key steps necessary for accurate problem-solving but with significantly reduced token counts. When mixed with the original long CoT reasoning examples, this combination allows student models to learn both comprehensive reasoning patterns and efficient reasoning shortcuts, resulting in models that can generate more concise reasoning during inference without sacrificing accuracy.

As illustrated in Figure \ref{fig:arch}, our approach can be divided into three distinct stages:
\textbf{(1) Structure-preserved CoT Rewriting}: We use a large language model as a rewriter model for the structure-preserved rewriting of long CoT trajectories. This process incorporates specific constraint instructions in the input prompt to ensure that the rewriting process maintains the logical structure and critical steps of the original reasoning path. Based on the existing long CoT reasoning dataset, this stage produces a corresponding short CoT reasoning dataset.
\textbf{(2) Mixture Supervised Fine-Tuning}: The original long CoT reasoning dataset and short CoT reasoning dataset obtained in the previous stage are completely randomly mixed to create a long-short mixture dataset. This mixed dataset is then used to perform supervised fine-tuning on a non-reasoning LLM.
\textbf{(3) Inference-time Balanced Thinking}: The mixture of both long and short CoT reasoning datasets enables student models to acquire comprehensive reasoning patterns while simultaneously learning efficient reasoning shortcuts. During inference, our model is provided with instructions that promote the generation of balanced thinking mode to solve the problem.

In the following subsections, we first provide a formal definition of the task, followed by a detailed explanation of each of the three stages of our method:

\subsection{Formal Task Definition}

Let $D_{\text{long}} = \{ (x_i, r^L_i, y_i) \}_{i=1}^{N}$ denote a long CoT reasoning dataset comprising $N$ instances, where $x_i$ represents a complex question, $r_i$ corresponds to the long CoT trajectory distilled from a LRM, and $y_i$ denotes the corresponding answer.

Our objective is to utilize this dataset through supervised fine-tuning to endow a non-reasoning LLM with effective reasoning capabilities.

\subsection{Structure-preserved CoT Rewriting}

A key component of our LS-Mixture SFT approach is the \textbf{structure-preserved CoT rewriting} methodology, which transforms verbose long CoT trajectories into more concise versions while preserving their core logical structure and key reasoning steps. This method significantly shortens the thinking part in the training data while preserving the reasoning process demonstrated by LRMs when addressing a problem, particularly the "aha moments" phenomenon exhibited by these reasoning-capable models.

We employ another large language model (Qwen2.5-72B-Instruct) as the rewriter model $\mathbb{P}_\text{rewriter}$, incorporating explicit constraints in the prompt template to instruct the model to maintain the original logical structure and critical steps of the CoT trajectory during rewriting. The prompt template employed by the rewriter model is presented in Appendix~\ref{subsec:appendix-rewriter-model-prompt}.

For each data point in the dataset $D_{\text{long}}$, we utilize the rewriter model to transform the long CoT trajectory $r^L_i$ into a shorter one $r^S_i$, which can be formally expressed as:

\begin{equation}
    r_i^S = \mathbb{P}_\text{rewriter}(r_i^L | x_i)
\end{equation}

After structure-preserved CoT rewriting, the short CoT trajectories are substantially shorter in length compared to their long CoT counterparts. Utilizing these rewritten short CoT trajectories, we are able to construct a short reasoning chain dataset $D_{\text{short}} = \{ (x_i, r_i^S, y_i) \}_{i=1}^{N}$.

\subsection{Mixture Supervised Fine-Tuning}

Following the previous stage that yields the short reasoning dataset $D_{\text{short}}$, we proceed to completely randomly merge it with the original long reasoning dataset $D_{\text{long}}$, creating a new mixed dataset $D_{\text{mix}}$:

\begin{equation}
    D_{\text{mix}} = D_{\text{long}} \cup D_{\text{short}}
\end{equation}

This mixture dataset $D_{\text{mix}}$ is then utilized to perform supervised fine-tuning on a non-reasoning LLM $M$ aiming to eliciting its efficient reasoning. To align with the current output format of LRMs, we encapsulate the CoT trajectory using special tokens \texttt{<think>} and \texttt{</think>}, and concatenate it with the answer part to form the ground-truth response for fine-tuning. The optimization objective $M^*$ can be formulated as follows:

\begin{equation}
\label{eq:LongCoTLoss}
    L(D_{\text{long}}) = \sum_{D_{\text{long}}} -\log \mathbb{P}_M( r^L_i \oplus y_i | x_i, p_\text{L})
\end{equation}

\begin{equation}
\label{eq:ShortCoTLoss}
    L(D_{\text{short}}) = \sum_{D_{\text{short}}} -\log \mathbb{P}_M( r^S_i \oplus y_i | x_i, p_\text{S})
\end{equation}

\begin{equation}
\label{eq:MixtureSFTObjective}
    M^* = \arg\min_{M} L(D_{\text{long}}) + L(D_{\text{short}})
\end{equation}

In Equations~\ref{eq:LongCoTLoss} and~\ref{eq:ShortCoTLoss}, $p_\text{L}$ and $p_\text{S}$ respectively represent the prompt that instruct the language model to reasoning in detailed and brief thinking modes. The specific prompt templates can be found in Appendix~\ref{subsec:appendix-detailed-thinking} and~\ref{subsec:appendix-brief-thinking}.

The mixture dataset ensures that the model is exposed to both comprehensive thinking patterns from long CoT trajectories and the efficient patterns from short ones, which enables the model to adapt its reasoning pattern based on the instruction type. When prompted with "detailed thinking" instructions, the model demonstrates comprehensive reasoning inherited from long CoT examples. Simultaneously, under "brief thinking" instructions, it employs concise yet effective reasoning patterns learned from short CoT examples.

\subsection{Inference-time Balanced Thinking}

Through our mixture training approach, the model simultaneously acquires both detailed and concise thinking modes. However, neither mode achieves an optimal balance between response effectiveness and efficiency. To address this limitation, we propose an inference-time balanced thinking methodology that leverages the dual reasoning capabilities developed during training while optimizing for both effectiveness and efficiency during model deployment.

To implement \textbf{balanced thinking mode}, we maintain the format of prompt template between the inference time and the training time, while modifying the instructions regarding the thinking mode. Specifically, we replace the directives for either detailed or brief thinking with instructions that encourage the model to engage in an "appropriate" thinking process that falls between these two extremes. This approach enables the model to balance effectiveness and efficiency in its reasoning process. The formulation can be expressed as follows:

\begin{equation}
    (r_i, y_i) = \mathbb{P}_{M^*}(x_i | p_\text{B})
\end{equation}

where $r_i$ is the approximate reasoning chain that is generated by the post-trained model $M^*$, and $p_\text{B}$ is the prompt template for balanced thinking. The specific prompt template can be found in Appendix~\ref{subsec:appendix-balanced-thinking}.

\section{Experiments}

\begin{table*}[!ht]
    \centering
    \caption{Results on 3 benchmarks. For each benchmark, we report both the response accuracy and response length in our evaluation results (with the exception of the o1 model). Due to accessibility limitations of the o1 model, we only report their publicly available scores on these benchmarks. Among these baseline models, s1.1-32B serves as our primary baseline model for comparison.}
    \label{tab:main-results}
    \begin{tabular}{cccccccc}
    \toprule
    \multirow{2}{*}[-\dimexpr\aboverulesep+\belowrulesep\relax]{\textbf{Model}} & \multicolumn{2}{c}{\textbf{MATH500}} & \multicolumn{2}{c}{\textbf{AIME24}} & \multicolumn{2}{c}{\textbf{GPQA}} & \multirow{2}{*}{\textbf{Avg. Length}} \\
    \cmidrule(lr){2-3} \cmidrule(lr){4-5} \cmidrule(lr){6-7}
    & Acc & Length & Acc & Length & Acc & Length & \\
    \midrule
    \multicolumn{8}{c}{\textbf{API only}} \\
    \midrule
    o1-previw & 85.5 & - & 44.6 & - & 73.3 & - & - \\
    o1 & 94.8 & - & 74.4 & - & 77.3 & - & - \\
    \midrule
    \multicolumn{8}{c}{\textbf{Open Weights}} \\
    \midrule
    DeepSeek R1 & 96.8 & 7,658.1 & 73.3 & 27,090.2 & 75.7 & 23,696.2 & 12820.9 \\
    QwQ-32B & 96.8 & 12,177.5 & 66.7 & 42,233.4 & 63.1 & 30,859.8 & 18,497.2 \\
    \midrule
    \multicolumn{8}{c}{\textbf{Open Weights and Open Data}} \\
    \midrule
    Sky-T1-32B & 85 & 6,839.1 & 50.0 & 7,893.9 & 53 & 10,376.5 & 7,844.6 \\
    \midrule
    s1.1-32B & $92.4_{\pm 1.6}$ & 12,351.4 & $53.3_{\pm 6.7}$ & 53,455.6 & $59.1_{\pm 2.0}$ & 56,040.7 & 25,927.8 \\
    \textbf{s1-mix-32B} & $\textbf{94.6}_{\pm 2.0}$ & $\textbf{8,648.7}$ & $\textbf{60.0}_{\pm 6.7}$ & $\textbf{40,251.3}$ & $\textbf{61.1}_{\pm 2.5}$ & \textbf{21,995.7} & $\textbf{13,581.1}_{\downarrow 47.6\%}$ \\
    \bottomrule
    \end{tabular}
\end{table*}

\subsection{Setup}   

\paragraph{Dataset} For experimental evaluation, we constructed \textbf{\data{}} using our mixture method. We utilized the s1K-1.1 dataset, which contains 1,000 instances of detailed reasoning trajectories and answers distilled from the DeepSeek R1 model, as our long CoT reasoning dataset ($D_{\text{long}}$). We implemented our structure-preserved CoT rewriting technique using Qwen2.5-72B-Instruct as the rewriter model. During rewriting process, 16 instances exceeded context length limitations, resulting in their exclusion from the dataset. The final short reasoning chain dataset ($D_{\text{short}}$) consisted of 984 examples. The mixture of these long and short examples constitutes our \textbf{\data{}} dataset. The statistics for the s1K-1.1 and s1K-mix datasets are presented in Appendix~\ref{sec:dataset-profile}.

\paragraph{Training} We perform supervised fine-tuning on Qwen2.5-32B-Instruct using the dataset \data{} to obtain our model \model{} using basic hyper parameters outline in Appendix~\ref{sec:training-hyperparameters}. All model training was conducted using the LlamaFactory \citep{LlamaFactory} framework. For all training samples, we use the delimiter \texttt{<think>} and \texttt{</think>} to separate the whole response into thinking part and the answering part. The relevant training hyper parameters (with the exception of the number of training epochs) are maintained consistent with those used for the \textbf{s1.1-32B} model \citep{s1-2501}. Given that the mixture dataset contains a greater number of samples than s1K-1.1, we adjusted the number of epochs to ensure that both models were exposed to an equivalent quantity of training samples. Let $N_{\text{long}}$ denote the number of training epochs for s1-32B ($N_{\text{long}} = 5$), $N_{\text{mix}}$ represent the number of training epochs for our \model{} model, and $|D_{\text{long}}|$ and $|D_{\text{mix}}|$ denote the size of the respective datasets. The numerical relationship is represented as: $N_{\text{long}} \times |D_{\text{long}}| = N_{\text{mix}} \times |D_{\text{mix}}|$.

\paragraph{Baselines} We benchmark \model{} against a series of top-tier models: OpenAI o1-series models: OpenAI o1 series \citep{OpenAI-o1}, representing close-source test-time scaling models; DeepSeek-R1 \citep{DeepSeek-R1} and QwQ-32B \citep{QwQ32b}, open-weight reasoning models; Sky-T1-32B \citep{SkyT1-2501} and \textbf{s1.1-32B} \citep{s1-2501}, open models with open reasoning data. Given that our pre-trained base model, training hyperparameters, and training data quantity are all consistent with those of the s1.1-32B model, the performance of the s1.1-32B model serves as the primary comparative baseline for our experiments. We have fully reproduced the training process of this model within our computational environment. Our models are fully open including weights, training dataset and code.

\paragraph{Benchmarks} We select three representative benchmarks widely used in the field: \textbf{MATH500} is a benchmark of competition math problems of varying difficulty. \textbf{AIME24} is a benchmark of high school level competition math problems. \textbf{GPQA Diamond} consists of 198 PhD-level science questions from Biology, Chemistry, and Physics. When we write "GPQA" in the context of our experiments, we refer to the GPQA Diamond subset. We evaluate the performance of models on these benchmarks using the LightEval \citep{lighteval} framework following the open-r1 project \citep{openr1}. In the experimental results, the presented accuracy corresponds to the median of five experiments.

\paragraph{Response Length Evaluation} In addition to evaluating accuracy on the benchmarks, we computed the average response length generated by models, including both the thinking part and the answer. This metric is crucial for assessing inference efficiency, as shorter responses directly translate to reduced latency and computational costs. We operate under the principle that, given comparable levels of accuracy, models that produce shorter responses are inherently more efficient and practical for real-world applications. For each model, we calculated the weighted average of response lengths across all benchmarks, using the number of samples in each evaluation dataset as weights for the weighted average computation.

\subsection{Results}   

Table~\ref{tab:main-results} presents the experimental results of our proposed \model{} model on the three benchmarks, highlighting the following key findings:

\textbf{(1) \model{} achieves a substantial reduction in model response length while imporving answer accuracy.} Despite utilizing the same training question set and equivalent number of training instances as s1.1-32B, our s1-mix-32B model attains accuracy improvements of 2.2\% on MATH500 (from 92.4\% to 94.6\%), 6.7\% on AIME24 (from 53.3\% to 60\%), and 2\% on GPQA (from 59.1\% to 61.1\%), all while reducing average response length by 47.61\% compared to s1.1-32B. These results demonstrate the effectiveness of our proposed method in enhancing both reasoning accuracy and efficiency.

\begin{figure}
    \centering
    \includegraphics[scale=0.4]{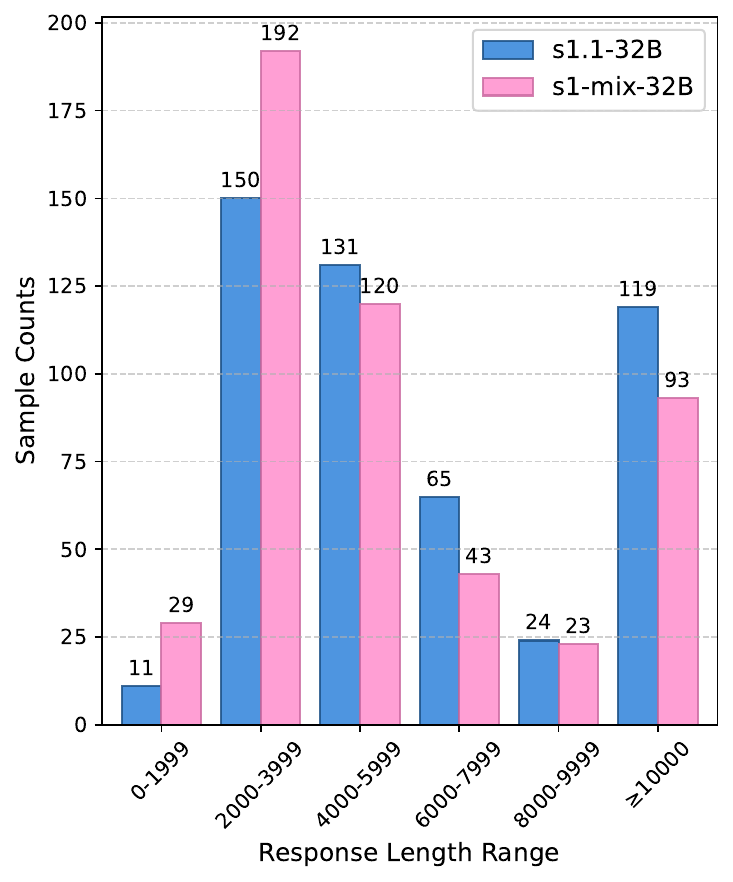}
    \caption{Comparison of response length distribution between \textbf{s1.1-32B} and \model{} models on the MATH500 evaluation task. The horizontal axis represents response length ranges (in string length), while the vertical axis shows the number of samples within each length range.}
    \label{fig:resp-len}
\end{figure}

\textbf{(2) The method proposed in this paper significantly reduces the computational cost of model training.} The \textbf{s1K-1.1} dataset used for training \textbf{s1.1-32B} has an average text length (including system prompt, question and model response) of 29,667.49 tokens. In contrast, our proposed mixture dataset reduces the average text length to 17,406.11 tokens—a reduction of 41.33\%. This decrease in average sequence length substantially lowers the training costs. Figure~\ref{fig:resp-len} presents the statistical analysis of response lengths generated by both models during evaluation.

\section{Ablations}  

\begin{table*}[h!]
    \centering
    \caption{Ablation experiment results on CoT rewriting strategies. We employed direct rewriting strategy and the ThinkTwice method to obtain short CoT trajectories. These datasets were then used to fine-tuning the Qwen2.5-32B-Instruct. We subsequently compared the task performance of models trained using different mixture datasets.}
    \label{tab:ablation-rewriting-results}
    \begin{tabular}{cccccccc}
    \toprule
    \multirow{2}{*}[-\dimexpr\aboverulesep+\belowrulesep\relax]{\textbf{Strategy}} & \multicolumn{2}{c}{\textbf{MATH500}} & \multicolumn{2}{c}{\textbf{AIME24}} & \multicolumn{2}{c}{\textbf{GPQA}} \\
    \cmidrule(lr){2-3} \cmidrule(lr){4-5} \cmidrule(lr){6-7}
    & Acc. & Length & Acc. & Length & Acc. & Length \\
    \midrule
    Direct & 85.4 & 6,106.1 & 33.3 & 5,876.4 & 53.5 & 31,204.0 \\
    ThinkTwice & 91 & 13,027.8 & 58.1 & 47,520.3 & 43.3 & 47,928.5 \\
    \textbf{Structure-preserved} & \textbf{94.6} & 8,648.7 & \textbf{60} & 40,251.3 & \textbf{61.1} & 21,995.7 \\
    \bottomrule
    \end{tabular}
\end{table*}

\begin{table*}[h!]
    \centering
    \caption{Ablation experiment results on dataset mixing strategies. We experimented with three datasets created by different strategies: only the long CoT reasoning dataset, only the short CoT reasoning dataset, and mixture dataset. We then evaluated the performance of the trained models across various tasks.}
    \label{tab:ablation-mixture-results}
    \begin{tabular}{cccccccc}
    \toprule
    \multirow{2}{*}[-\dimexpr\aboverulesep+\belowrulesep\relax]{\textbf{Mix Method}} & \multicolumn{2}{c}{\textbf{MATH500}} & \multicolumn{2}{c}{\textbf{AIME24}} & \multicolumn{2}{c}{\textbf{GPQA}} \\
    \cmidrule(lr){2-3} \cmidrule(lr){4-5} \cmidrule(lr){6-7}
    & Acc. & Length & Acc. & Length & Acc. & Length \\
    \midrule
    Long-only & 92.4 & 12,351.4 & 53.3 & 53,455.6 & 59.1 & 56,040.7 \\
    Short-only & 82.6 & 3,205.8 & 16.7 & 6,646.3 & 49.0 & 3,961.7 \\
    \textbf{Mixture} & \textbf{94.6} & 8,648.7 & \textbf{60} & 40,251.3 & \textbf{61.1} & 21,995.7 \\
    \bottomrule
    \end{tabular}
\end{table*}

\begin{table*}[h!]
    \centering
    \caption{Ablation experiment results on different thinking modes. We evaluated the performance of the \model{} model across various tasks using three thinking modes: detailed thinking, brief thinking, and balanced thinking.}
    \label{tab:ablation-thinking-results}
    \begin{tabular}{cccccccc}
    \toprule
    \multirow{2}{*}[-\dimexpr\aboverulesep+\belowrulesep\relax]{\textbf{Thinking Mode}} & \multicolumn{2}{c}{\textbf{MATH500}} & \multicolumn{2}{c}{\textbf{AIME24}} & \multicolumn{2}{c}{\textbf{GPQA}} \\
    \cmidrule(lr){2-3} \cmidrule(lr){4-5} \cmidrule(lr){6-7}
    & Acc. & Length & Acc. & Length & Acc. & Length \\
    \midrule
    Brief & 81.0 & 2,963.9 & 20.0 & 4,490.3 & 52.0 & 4,125.0 \\
    Detailed & 92.6 & 1,162.3 & 56.7 & 53,107.4 & \textbf{62.1} & 41,762.9 \\
    \textbf{Balanced} & \textbf{94.6} & 8,648.7 & \textbf{60} & 40,251.3 & 61.1 & 21,995.7 \\
    \bottomrule
    \end{tabular}
\end{table*}

\subsection{Impact of Rewriting Strategies}
\label{subsec:ablations-rewriting-strategy}

To investigate the importance of preserving reasoning logical structure during the rewriting CoT from long to short chains, we conducted ablation studies comparing three distinct rewriting strategies: \textbf{Direct compression}: A straightforward approach where the LLM is instructed to compress the long reasoning chain freely. The specific prompt templates can be found in Appendix~\ref{subsec:appendix-direct-compression-prompt}. \textbf{ThinkTwice method}: Inspired by the ThinkTwice method \citep{ThinkTwice-2503}, this approach incorporates the answer into the specific prompt used to model generation. The thinking part produced during generation serve as the shortened CoT. \textbf{Structure-preserved rewriting} (Ours): Our proposed approach that explicitly instructs the rewriter model to maintain the reasoning logical structure and critical steps while performing rewriting.

For each rewriting strategy, we created a corresponding short reasoning dataset and applied our method to fine-tune Qwen2.5-32B-Instruct. As demonstrated in the Table~\ref{tab:ablation-rewriting-results}, all alternative chain-of-thought rewriting methods resulted in diminished model training effectiveness, highlighting the importance of preserving the original logical reasoning structure during rewriting stage.

\subsection{Impact of Long-Short Chain Mixing Strategies}

To investigate the effectiveness of our proposed mixing strategy, we conducted experiments comparing three distinct training datasets. \textbf{Long-only}: The thinking part of the data point exclusively comprises long CoT trajectories, specifically $D_{\text{long}}$. \textbf{Short-only}: The thinking part exclusively comprises short CoT trajectories, specifically $D_{\text{short}}$. \textbf{Mixture}: The dataset obtained using the mixture method proposed in this paper, namely $D_{\text{mix}}$. The training configurations employed in these experiments are consistent with those utilized in our primary experiments.

Table~\ref{tab:ablation-mixture-results} presents the experimental results across our evaluation benchmarks. The results demonstrate that our proposed mixing strategy consistently outperforms other approaches.

\subsection{Impact of Inference-time Thinking Modes}

During the training of \model{}, detailed and brief thinking modes were employed for the long and short CoT reasoning dataset, respectively, while a balanced thinking mode was utilized for problem-solving during inference. To investigate the impact of different thinking modes at inference time, we conducted evaluations using these three distinct thinking modes for \model{}. As shown in Table~\ref{tab:ablation-thinking-results}, employing the balanced thinking mode yields the optimal results, which validate our hypothesis that the balanced thinking mode during inference time can effectively leverage both the comprehensive thinking capabilities learned from long CoT examples and the efficient reasoning patterns acquired from short counterparts.

\section{Discussion}

\subsection{Structure-preserving in CoT Rewriting}

Our findings highlight the importance of maintaining the core logical structure when rewriting long CoT trajectories into short formats. Our ablation studies~\ref{subsec:ablations-rewriting-strategy}, which explore various strategies for CoT trajectory rewriting, revealed a insight: overly simplified CoT fail to adequately stimulate the student model's reasoning capabilities. Conversely, we observed that by preserving the original core structure and key steps from the long CoT trajectories during the rewriting stage could the student model be guided to learn how to reason effectively. Notably, models fine-tuned under these conditions exhibited "aha moments" phenomenon \citep{DeepSeek-R1} which was observed primarily in models trained through reinforcement learning.

\subsection{Analysis of Response Length in Incorrect Answers}

\begin{figure}
  \centering
  \includegraphics[scale=0.3]{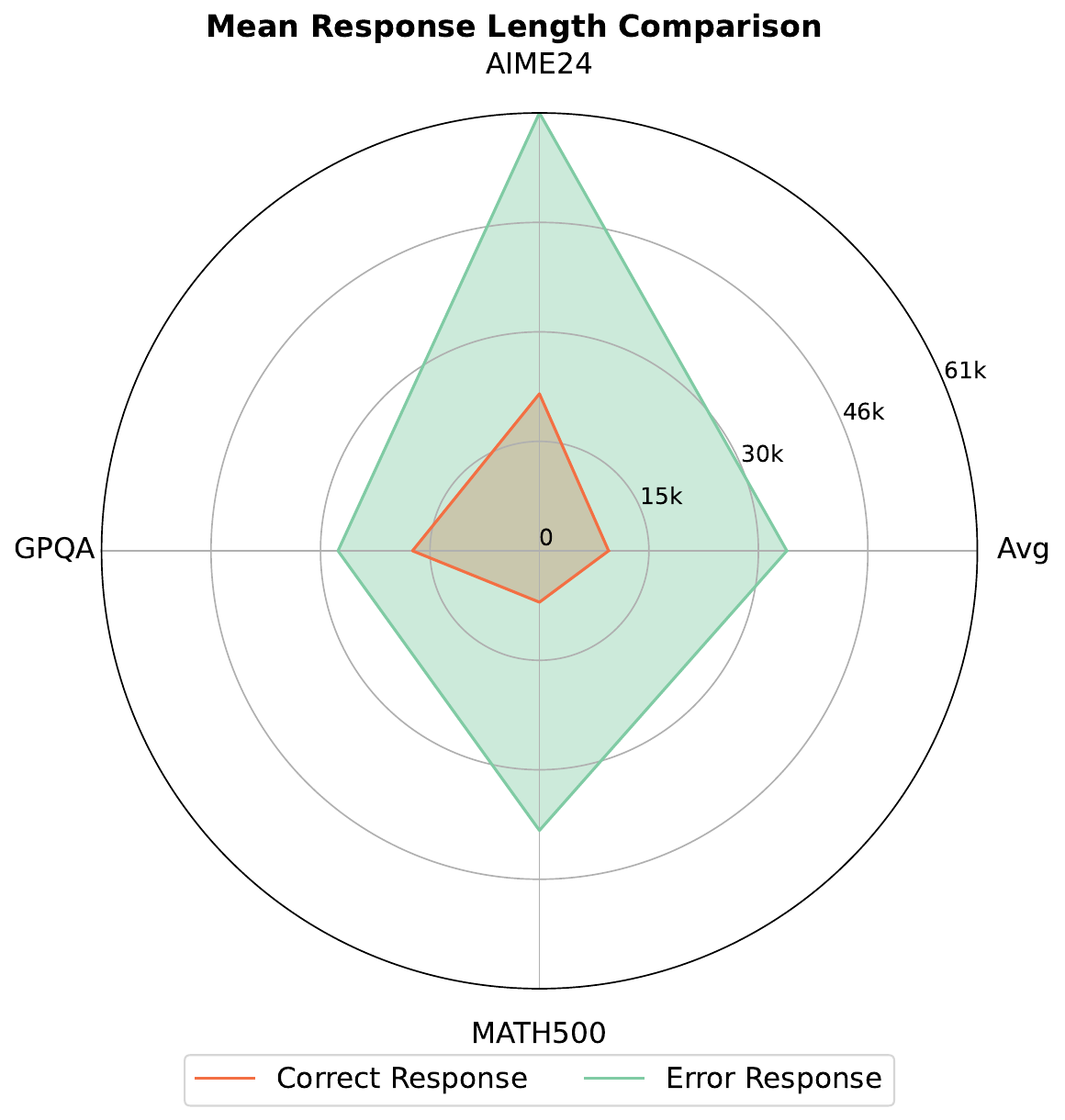}
  \caption{Comparison of mean response lengths for correct (red) and error (green) predictions of the \model{} model across the MATH500, GPQA, and AIME24 evaluation datasets. The figure also presents a weighted average of these results, with weights determined by the size of each respective dataset.}
  \label{fig:mean-length-comparison}
\end{figure}

Our analysis of the evaluation results from the \model{} reveals a correlation between response correctness and length. As illustrated in Figure~\ref{fig:mean-length-comparison}, incorrect responses demonstrate greater verbosity across all evaluation datasets. Specifically, the average response length of incorrect examples is approximately 3.58 times longer than the correct part. This finding suggests that when the model is uncertain or unable to produce an accurate answer, it tends to generate more unhelpful text.

\section{Related Work}

\subsection{Chain-of-Thought Reasoning in LLMs}

Chain-of-Thought reasoning has emerged as a pivotal technique for enhancing the reasoning capabilities of large language models. Initially introduced by \citet{CoT-2022}, CoT prompting encourages models to generate intermediate reasoning steps before producing a final answer \citep{ARES}. This approach has proven particularly effectively for complex reasoning tasks \citep{Li2024-CoT, Madaan2023-CoT}, including mathematical problem-sovling \citep{MuMath}, logical reasoning \citep{logicasker,Toroghi2024-Verifiable}, and scientific inquery \citep{SciEval}.

With the recent discovery of test-time scaling laws \citep{Wu2024-TestTimeScaling, Snell2025-TestTimeScaling}, Large Reasoning Models, exemplified by DeepSeek R1 \citep{DeepSeek-R1}, have undergone substantial development. These works utilize techniques such as reinforcement learning to enable LLMs to generate a CoT reasoning process enclosed by special tokens (e.g., \texttt{<think>} and \texttt{</think>}) \citep{O1ReplicationJourney-Part1, Kimi-K1.5, LightR1}.

The emergence of LRMs has further enhanced the capabilities of LLM on complex reasoning tasks. However, to transfer these reasoning abilities to non-reasoning models such as Qwen-2.5 series \citep{Qwen2.5-TechnicalReport}, current research \citep{Zhang2025-100DaysDeepSeekR1Survey, s1-2501} has found that it is also possible to elicit thier reasoning abilities by performing supervised fine-tuning on non-reasoning models using dataset distilled from LRMs \citep{Zhang2025-TestTimeScalingSurvey, Chen2025-EmpiricalStudyOnEliciting}. Our approach follows this line of research, refining the fine-tuning methodology to elicit efficient reasoning capabilities in non-reasoning models.

\subsection{Efficient Reasoning in LRMs}

While LRMs improve performance in System-2 reasoning domains \citep{Li2025-FromSystem1toSystem2}, they also introduce significant computational overheads due to verbose and redundant reasoning steps, known as the "overthinking phenomenon" \citep{SurveyEfficientReasoning-2503, StopOverthinkingSurvey-2303}. To address this issue, a series of efficient reasoning \citep{Feng2025-EfficientReasoningSurvey, ChainOfDraft, TokenComplexity, ThoughtManipulation} methods have been proposed to enhance the inference-time efficiency of LRMs. These methods vary in approach: some incorporate response length-related rewards into reinforcement learning \citep{L1, O1Pruner, DAST, Demystifying}, others differentiate problem difficulty levels to allocate token budgets accordingly \citep{RoouteLLM, SketchOfThought, Self-Calibration}, and yet others leverage smaller models to achieve faster thinking processes \citep{SplitReason, SCoT}. To our knowledge, our approach inspired by C3oT \citep{C3oT} is the first investigation from the supervised fine-tuning perspective on achieving efficient reasoning goals while eliciting reasoning capabilities in non-reasoning models through distillation from LRMs.

\section{Conclusion}

We presented \textbf{LS-Mixture SFT}, a novel approach for eliciting efficient reasoning capabilities in non-reasoning models using dataset distilled from large reasoning models, thereby enabling the trained model to maintain task performance while reducing response length, and avoiding the inheritance of the overthinking problem from teacher models to student models during the distillation process. Based on our proposed method, we constructed the \data{} dataset and the \model{} model. Through experiments conducted across multiple benchmarks, we find that compared to the baseline model, our model achieves consistently higher accuracy while significantly reducing response length.

In future work, we will continue to explore how our method can be integrated with token-level text compression to further reduce redundant information in thinking trajectories and enhance model performance on relevant tasks.

\section*{Limitations}

Despite the promising results presented in this paper, our study is subject to several limitations. The experiments conducted in this work were restricted to a 32B parameter model and datasets containing only 1,000 examples. Due to computational resource constraints, we were unable to extend our experiments to larger-scale models or more extensive datasets, which could potentially reveal different scaling behaviors or effects.

Furthermore, we did not thoroughly investigate the impact of varying mixture ratios between long and short Chain-of-Thought trajectories during training. The optimal balance between these different types of reasoning demonstrations may vary across different model sizes, tasks, and domains. This represents an important dimension for future exploration that could yield further improvements in model performance and efficiency.

\section*{Ethics Statement}

This research utilizes the s1K-1.1 dataset and Qwen2.5 series models, both of which are publicly available online resources. We have provided appropriate citations to acknowledge the original work behind these resources. Our study focuses on improving model training efficiency through Chain-of-Thought trajectory rewriting techniques, which does not introduce new ethical concerns beyond those inherent to large language model research.

\bibliography{latex/main}

\appendix
\section{Prompt Template}
\label{sec:appendix-prompt-template}

\subsection{The prompt of Rewriter Model}
\label{subsec:appendix-rewriter-model-prompt}

\begin{tcolorbox}[title=Rewriter Model,label=box:rewrite-model-prompt]
You have a QUESTION and a THOUGHT PROCESS now, and you need to simplify the THOUGHT PROCESS while maintaining its original structure and steps. \\

QUESTION: \{question\} \\

THOUGHT PROCESS: \{thought\_process\} \\

Now, you need to simplify the THOUGHT PROCESS while maintaining its original structure and steps. For each step in the original THOUGHT PROCESS:\\
1. Keep the original logical flow and steps as much as possible, including the thinking process, verification process, and the final answer.\\
2. Remove redundant tokens.\\
3. Preserve the step-by-step format.\\
4. Allow condensed thought processes to include attempts at different reasoning processes.\\
Do not add any new information that wasn't in the original THOUGHT PROCESS.\\

SIMPLIFIED THOUGHT PROCESS:
\end{tcolorbox}

\subsection{The prompt of detailed thinking mode}
\label{subsec:appendix-detailed-thinking}

\begin{tcolorbox}[title=Detail Thinking Mode]
Answer the problem with a \textbf{detailed} thinking process:
\end{tcolorbox}

\subsection{The prompt of brief thinking mode}
\label{subsec:appendix-brief-thinking}

\begin{tcolorbox}[title=Brief Thinking Mode]
Answer the problem with a \textbf{brief} thinking process:
\end{tcolorbox}

\subsection{The prompt of balanced thinking mode}
\label{subsec:appendix-balanced-thinking}

\begin{tcolorbox}[title=Balanced Thinking Mode]
Answer the problem with a \textbf{appropriate} (between detailed and brief) thinking process:
\end{tcolorbox}

\subsection{The prompt of Direct Compression method}
\label{subsec:appendix-direct-compression-prompt}

\begin{tcolorbox}[title=Direct Compression]
You have a question now:\\

QUESTION:\\
\{question\}\\

THOUGHT PROCESS:\\
\{thought\_process\}\\

Now, you need to simplify the THOUGHT PROCESS as short as possible to only include the key information needed to solve the question. And do not add additional information that is not included in the original THOUGHT PROCESS.\\

SIMPLIFIED THOUGHT PROCESS:
\end{tcolorbox}

\section{Dataset Profile}
\label{sec:dataset-profile}

\begin{table}[htbp] 
\centering 
\caption{The statistical profile of the datasets used in this study, namely s1K-1.1 and s1K-mix. For each dataset, we report the number of rows and the average text length.}
\label{tab:dataset_stats} 
\begin{tabular}{@{}ccc@{}} 
\toprule 
\textbf{Dataset} & \textbf{Num of Rows} & \textbf{Average Length} \\ 
\midrule 
s1K-1.1 & 1000 & 29667.49 \\
s1K-mix & 1984 & 17406.11 \\
\bottomrule 
\end{tabular}
\end{table}

\section{Training Hyperparameters}
\label{sec:training-hyperparameters}

All experiments were run in a GPU cluster of 16 * A800. The hyperparameters used for training are presented in Table \ref{tab:hyperparams}, while any parameters not explicitly specified utilize the default values provided by LlamaFactory \citep{LlamaFactory}.

\begin{table}[h!]
    \centering
    \caption{Training Hyperparameters}
    \begin{tabular}{lc}
    \toprule
    \textbf{Hyperparameter} & \textbf{Value} \\
    \midrule
    cutoff\_len & 4096 \\ 
    learning\_rate & 1e-5 \\
    lr\_scheduler\_type & cosine \\
    warmup\_ratio & 0.05 \\
    bf16 & true \\
    optimizer & AdamW \\
    weight\_decay & 1e-4 \\
    \bottomrule
    \end{tabular}
    \label{tab:hyperparams}
\end{table}


\section{Word cloud of Datasets}
\label{sec:wordcloud}

\begin{figure}[H]
    \centering
    \includegraphics[scale=0.2]{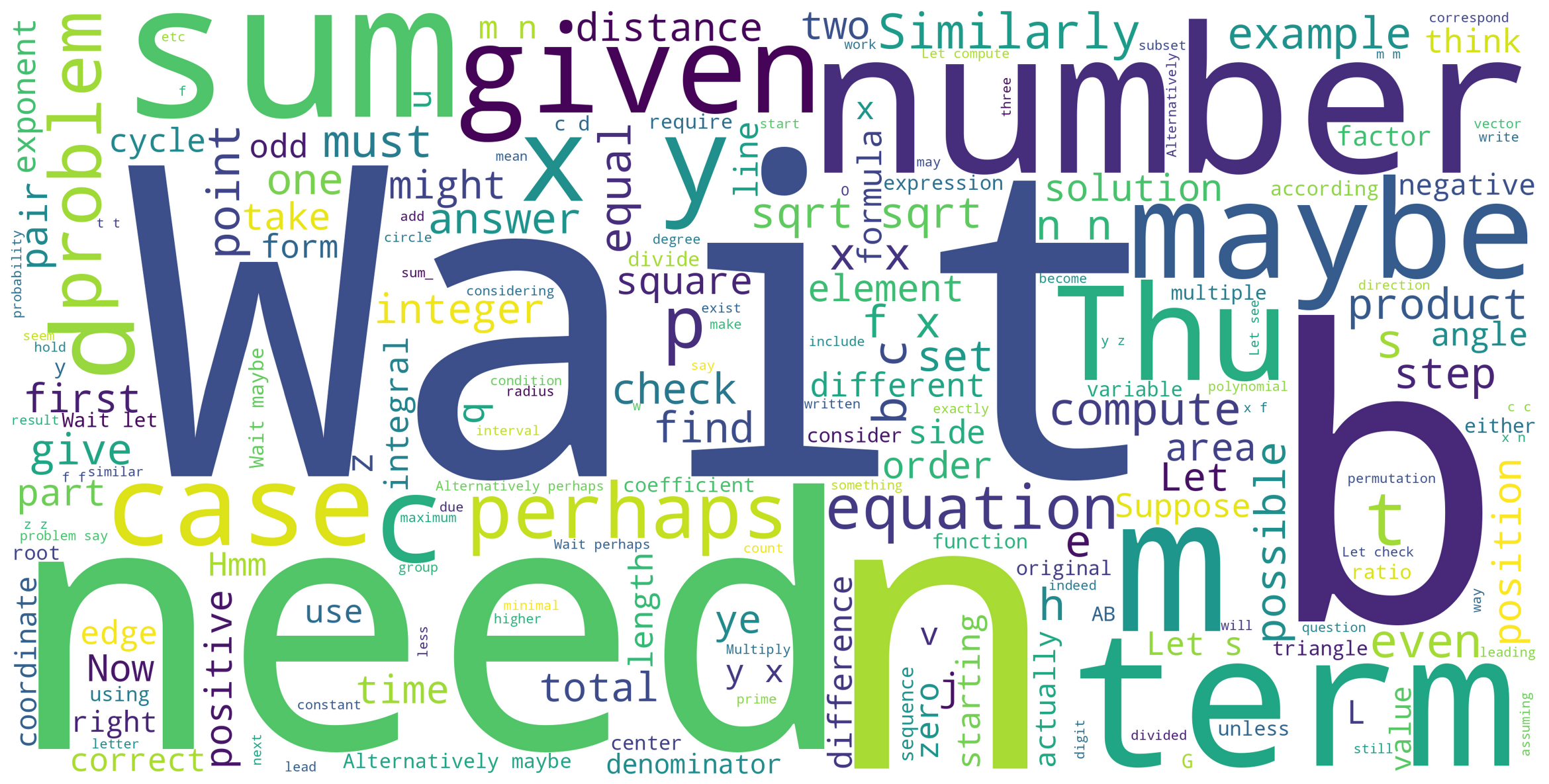}
    \caption{Word cloud of CoT trajectories in the long reasoning dataset.}
    \label{fig:long-cot-wordcloud}
\end{figure}

\begin{figure}[H]
    \centering
    \includegraphics[scale=0.2]{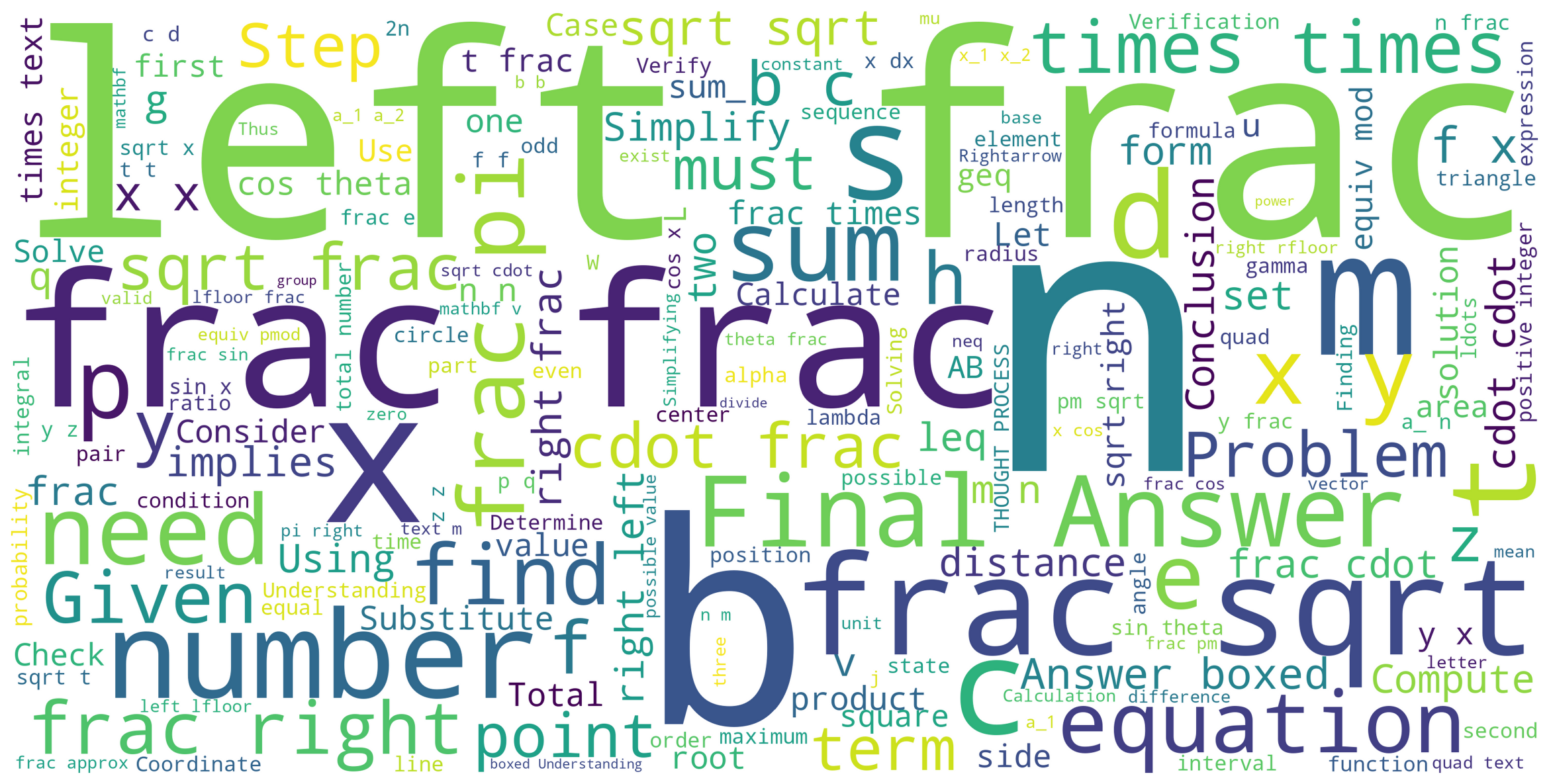}
    \caption{Word cloud of CoT trajectories in the short reasoning dataset.}
    \label{fig:short-cot-wordcloud}
\end{figure}

Figures~\ref{fig:long-cot-wordcloud} and \ref{fig:short-cot-wordcloud} respectively display word clouds of the chain-of-thought trajectories from our experiments on the long reasoning dataset and the short reasoning dataset. As can be observed from these two figures, the distribution of common words undergoes a significant change following structure-preserved rewriting, notably marked by the disappearance of the words 'wait' and 'need'.

\end{document}